\documentclass{article}
\usepackage{stywhispers}
\usepackage{times}
\usepackage{epsfig}
\usepackage{graphicx}
\usepackage{amsmath}
\usepackage{amssymb}
\usepackage{graphicx}
\graphicspath{{./pics/}}
\usepackage[usenames, dvipsnames]{color}
\usepackage{amsthm}
\usepackage{amsfonts}
\usepackage{amscd}
\usepackage{mathrsfs}
\usepackage{bm}
\usepackage{enumerate}
\usepackage{algorithmic, algorithm}
\usepackage{multirow}
\usepackage{hyperref}
\newcommand{\R}{\mathbb{R}}
\newcommand{\M}{\mathcal{M}}

\newcommand{\CP}{\mathcal{P}}
\newcommand{\bb}{\bm{b}}
\newcommand{\bu}{\bm{u}}
\newcommand{\bx}{\bm{x}}

\newcommand{\by}{\bm{y}}
\newcommand{\bp}{\bm{p}}

\newtheorem{remark}{Remark}
\usepackage{algorithmic, algorithm}
\usepackage{array}
\newcolumntype{C}[1]{>{\centering\let\newline\\\arraybackslash\hspace{0pt}}m{#1}}

\floatname{algorithm}{Procedure}

\title{Scalable low dimensional manifold model in the reconstruction of noisy and incomplete hyperspectral images}
\threeauthors
  {Wei Zhu}
	{Duke University\\
	Mathematics Department\\\\
	\tt{zhu@math.duke.edu}}
  {Zuoqiang Shi\sthanks{Equal contribution. This work is supported by NSFC: 11671005.}}
	{Tsinghua University\\
	Department of Mathematical Sciences\\ Yau Mathematical Sciences Center\\
	\tt{zqshi@tsinghua.edu.cn}}
  {Stanley Osher\sthanks{This work is supported by NSF: DMS-1737770, and STROBE: DMR 1548924.}}
        {UCLA\\
	Department of Mathematics\\\\
	\tt{sjo@math.ucla.edu}}

\begin{document}
%
\maketitle
\begin{abstract}
We present a scalable low dimensional manifold model for the reconstruction of noisy and incomplete hyperspectral images. The model is based on the observation that the spatial-spectral blocks of a hyperspectral image typically lie close to a collection of low dimensional manifolds. To emphasize this, the dimension of the manifold is directly used as a regularizer in a variational functional, which is solved efficiently by alternating direction of minimization and weighted nonlocal Laplacian. Unlike general 3D images, the same similarity matrix can be  shared across all spectral bands for a hyperspectral image, therefore the resulting algorithm is much more scalable than that for general 3D data \cite{ldmm_scientific}. Numerical experiments on the reconstruction of hyperspectral images from sparse and noisy sampling demonstrate the superiority of our proposed algorithm in terms of both speed and accuracy.
\end{abstract}
\section{Introduction}
\label{sec:intro}

A hyperspectral image (HSI) is a collection of 2D images of the same spatial location taken at hundreds of different wavelengths \cite{chang2003hyperspectral}. The observed images are typically degraded when such data of high dimensionality are collected. For instance, the images can be very noisy due to limited exposure time, or some of the voxels can be missing due to the malfunctions of the hyperspectral cameras. An important task in HSI analysis is to recover the original image from its noisy incomplete observation. This is an ill-posed inverse problem, and some prior knowledge of the original data must be exploited.

One widely used prior information of HSI is that the 3D data cube has a low-rank structure under the linear mixing model (LMM) \cite{unmixingoverview}. More specifically, the spectral signature of each pixel is assumed to be a linear combination of a few constituent endmembers. Under such an assumption, low-rank matrix completion and sparse representation techniques have been used for HSI reconstruction \cite{sparse_hsi_1,sparse_hsi_2,duke_inpaint}. Despite the simplicity of LMM, the linear mixing assumption has been shown to be physically inaccurate in certain situations \cite{nonlinearunmixing}.

Various partial differential equation (PDE) and graph based image processing techniques have also been applied to HSI reconstruction. The total variation (TV) method \cite{ROF92} has been widely used as a regularization in hyperspectral image processing \cite{Yuan2012, Iordache2012, He2016, Aggarwal2016}. The nonlocal total variation (NLTV) \cite{oshernonlocal}, which computes the gradient in a nonlocal graph-based manner, has also been applied to the analysis of hyperspectral images \cite{huiyi_plume, zhu_primaldual, Li2015}. However, such methods fail to produce satisfactory results when there is a significant number of missing voxels.

In \cite{ldmm, ldmm_wgl}, the authors proposed a low dimensional manifold model (LDMM) for general image processing problems. LDMM is based on the observation that patches of a natural image typically sample a collection of low dimensional manifolds. Therefore the dimension of the patch manifold is directly used as a regularization term in a variational functional. The resulting Euler-Lagrange equation is solved either by the point integral method (PIM) \cite{SS_neumann}, or the weighted nonlocal Laplacian \cite{wgl}. LDMM achieved excellent results, especially in image inpainting problems from very sparse sampling. LDMM was also extended to 3D scientific data interpolation \cite{ldmm_scientific}, but such generalization has poor scalability and requires huge memory storage.

In this paper, we exploit the special structure of hyperspectral images and propose a scalable LDMM specifically designed for the reconstruction of HSI from noisy and sparse sampling. The rationale behind the proposed method is that a hyperspectral image is a collection of 2D images of the same spatial location, and hence a single spatial similarity matrix can be shared across all spectral bands. The resulting algorithm is considerably faster than its 3D counterpart: it typically takes less than two minutes given a proper initialization as compared to hours in \cite{ldmm_scientific}.

\section{LDMM for HSI reconstruction}
\subsection{Patch Manifold}
We first describe the patch manifold of a hyperspectral image.  Let $\bu \in \mathbb{R}^{m \times n \times B}$ be a hyperspectral image, 
where $m \times n$ and $B$ are the spatial and spectral dimensions of the image. 
For any $\bx\in \bar{\Omega}=[m]\times [n]$, where $[m]= \{1,2,\ldots,m\}$, 
we define a patch $\CP_{\bx}(\bu)$ as a 3D block of size $s_1\times s_2 \times B$ of the original data cube $\bu$, 
and the pixel $\bx$ is the top-left corner of the rectangle of size $s_1\times s_2$. The \textit{patch set} $\mathcal{P}(\bu)$ is defined as the collection of all patches:
\begin{equation}
	\mathcal{P}(\bu)=\{\CP_{\bx}(\bu): \bx\in \bar{\Omega} \} \subset \mathbb{R}^d,\quad d=s_1\times s_2\times B.
\end{equation}

Previous work \cite{ldmm_scientific, ldmm} has shown that the point cloud $\CP(\bu)$ is typically close to a collection of low dimensional smooth manifolds $\mathcal{M} = \cup_{l=1}^L\M_l$ embedded in $\mathbb{R}^{d}$. This collection of manifolds is called the {\it patch manifold} of $\bu$.

\subsection{Scalable LDMM}

Our objective is to reconstruct the unknown HSI $\bu$ from its noisy and incomplete observation $\bb \in \mathbb{R}^{m \times n \times B}$. Assume that for each spectral band $t\in [B]$, $\bb$ is only known on a random subset $\Omega^{t}\subset \bar{\Omega}$, with a sampling rate $r$ (in our experiments $r=5\%$ or $10\%$). According to \cite{ldmm_scientific,ldmm}, we can use the dimension of the patch manifold as a regularizer to reconstruct $\bu$ from $\bb$:
\begin{align}
  \nonumber
  &\min_{\bu\in \mathbb{R}^{m\times n \times B}\atop \M\subset \mathbb{R}^d} \int_\M\dim(\mathcal{M}(\bm{p}))d\bm{p}+ \lambda\sum_{t=1}^B \|\bu^t-\bb^t\|_{L^2(\Omega^t)^2}\\
  \label{eq:manifold-model}
  &\text{subject to:}\quad \CP(\bu)\subset \M,
\end{align}
where $\bu^t$ is the $t$-th spectral band of the HSI $\bu$, $\M(\bm{p})$ denotes the smooth manifold $\M_l$ to which $\bm{p}$ belongs, and $\int_\M\dim(\mathcal{M}(\bm{p}))d\bm{p}=\sum_{l=1}^L|\M_l|\dim(\M_l)$ is the $L^1$ norm of the local dimension. Based on Proposition 3.1 in \cite{ldmm}, the first term in (\ref{eq:manifold-model}) can be written as the $L^2$ norm of the coordinate functions $\alpha_i^t: \M \rightarrow \R$. More specifically, (\ref{eq:manifold-model}) is equivalent to
\begin{align}
  \nonumber
  &\min_{\bu\in \mathbb{R}^{m\times n \times B}\atop \M\subset \mathbb{R}^d} \sum_{i=1}^{d_s}\sum_{t=1}^B\|\nabla_\M \alpha_i^t\|_{L^2(\M)}^2+ \lambda\sum_{t=1}^B \|\bu^t-\bb^t\|_{L^2(\Omega^t)^2}\\
  \label{eq:model-manifold-grad}
  &\text{subject to:}\quad \CP(\bu)\subset \M,
\end{align}
where $d_s= s_1\times s_2$ is the spatial dimension, $\alpha_i^t$ is the coordinate function that maps every point $\bp = \left(p_i^t\right)_{i,t}\in \M$ into its $(i,t)$-th coordinate $p_i^t$. Note that (\ref{eq:manifold-model}) is nonconvex, and we solve it by alternating the direction of minimization with respect to $\bu$ and $\M$. More specifically, given $\M^{(k)}$ and $\bu^{(k)}$ at step $k$ satisfying $\CP(\bu^{(k)}) \subset \M^{(k)}$:
\begin{itemize}
\item With fixed $\M^{(k)}$, update the data $\bu^{(k+1)}$ by solving:
\begin{align}
 \nonumber
  &\min_{\bu}  \sum_{i,t}\|\nabla_{\M^{(k)}} \alpha_i^t\|_{L^2(\M^{(k)})}^2+\lambda\sum_{t=1}^B \|\bu^t-\bb^t\|_{L^2(\Omega^t)}^2\\  \label{eq:update-u}
&\text{subject to: } \alpha_i^t(\CP\bu^{(k)}(\bx))=\CP_i^t \bu(\bx),\quad \bx\in \overline{\Omega}
\end{align}
where $\CP_i^t \bu(\bx)$ is the $(i,t)$-th element in the patch $\CP_{\bx}\bu$.
\item Update the manifold $\M^{(k+1)}$ as the image under the perturbed coordinate function $\bm{\alpha}$:
  \begin{align}
    \label{eq:update-manifold}
    \M^{(k+1)} = \bm{\alpha}(\M^{(k)})
  \end{align}
\end{itemize}
The manifold update (\ref{eq:update-manifold}) is easy to implement, and \cite{ldmm_wgl,ldmm_scientific} introduced a way to solve (\ref{eq:update-u}) using the weighted nonlocal Laplacian (WNLL) \cite{wgl}. The idea is to  discretize the Dirichlet energy $\|\nabla_{\M^{(k)}} \alpha_i^t\|_{L^2(\M^{(k)}}^2$  as
\begin{align}\nonumber
  &\frac{|\bar{\Omega}|}{|\Omega_i^t|}\sum_{\bx \in \Omega_i^t}\sum_{\by \in \bar{\Omega}}\bar{w}(\bx,\by)\left(\alpha_i^t(\CP\bu^{(k)}(\bx))-\alpha_i^t(\CP\bu^{(k)}(\by))\right)^2\\ \label{eq:wgl}
  + & \sum_{\bx \in \bar{\Omega}\setminus\Omega_i^t}\sum_{\by \in \bar{\Omega}}\bar{w}(\bx,\by)\left(\alpha_i^t(\CP\bu^{(k)}(\bx))-\alpha_i^t(\CP\bu^{(k)}(\by))\right)^2,
\end{align}
where $\Omega_i^t = \left\{\bx \in \bar{\Omega}: \CP_i^t\bu^{(k)}(\bx) \text{ is sampled}\right\}$ is a spatially translated version of $\Omega^t$, $|\bar{\Omega}|/|\Omega_i^t| = 1/r$ is the inverse of the sampling rate, and $\bar{w}(\bx,\by) = w(\CP\bu^{(k)}(\bx),\CP\bu^{(k)}(\by))$ is the similarity between the patches, with
\begin{align}
  \label{eq:weight}
  w(\bm{p},\bm{q}) = \exp \left(-\frac{\|\bm{p}-\bm{q}\|^2}{\sigma(\bm{p})\sigma(\bm{q})}\right),
\end{align}
where $\sigma(\bm{p})$ is the normalizing factor. Combining the WNLL discretization (\ref{eq:wgl}) and the constraint in (\ref{eq:update-u}), the update of $\bu$ in (\ref{eq:update-u}) can be discretized as
\begin{align}\nonumber
  &\min_{\bu} \quad \lambda\sum_{t=1}^B \|\bu^t-\bb^t\|_{L^2(\Omega^t)}^2\\ \nonumber
  &+\sum_{i,t}\left[\sum_{\bx \in \bar{\Omega}\setminus\Omega_i^t}\sum_{\by \in \bar{\Omega}}\bar{w}(\bx,\by)\left(\CP_i^t\bu(\bx)-\CP_i^t\bu(\by)\right)^2\right.\\\label{eq:update-u-wgl}
  & +\left.\frac{1}{r}\sum_{\bx \in \Omega_i^t}\sum_{\by \in \bar{\Omega}}\bar{w}(\bx,\by)\left(\CP_i^t\bu(\bx)-\CP_i^t\bu(\by)\right)^2\right].
\end{align}
\begin{remark}
  Unlike the model in \cite{ldmm_scientific}, the similarity matrix $\bar{w}$ in (\ref{eq:update-u-wgl}) is built on 2D coordinates $\bx,\by \in \bar{\Omega}$, which significantly improves the scalability of the model.
\end{remark}
Note that (\ref{eq:update-u-wgl}) is decoupled with respect to the spectral coordinate $t$, and for any given $t\in [B]$, we only need to solve the following problem:
\begin{align}\nonumber
  &\min_{\bu^t} \quad \lambda\|\bu^t-\bb^t\|_{L^2(\Omega^t)}^2\\ \nonumber
  &+\sum_{i=1}^{d_s}\left[\sum_{\bx \in \bar{\Omega}\setminus\Omega_i^t}\sum_{\by \in \bar{\Omega}}\bar{w}(\bx,\by)\left(\CP_i\bu^t(\bx)-\CP_i\bu^t(\by)\right)^2\right.\\\label{eq:update-u-decouple}
  & +\left.\frac{1}{r}\sum_{\bx \in \Omega_i^t}\sum_{\by \in \bar{\Omega}}\bar{w}(\bx,\by)\left(\CP_i\bu^t(\bx)-\CP_i\bu^t(\by)\right)^2\right].
\end{align}
A standard variational technique shows that (\ref{eq:update-u-decouple}) is equivalent to the following Euler-Lagrange equation:
\begin{align}\nonumber
  0  =  &\mu \sum_{i=1}^{d_s}\CP_i^* I_{\Omega_i^t}\left[\sum_{\by \in \bar{\Omega}}\bar{w}(\bx,\by)\left(\CP_i \bu^t(\bx)-\CP_i \bu^t(\by)\right)\right]\\\nonumber
  &+\sum_{i=1}^{d_s}\CP_i^*\left[\sum_{\by \in \bar{\Omega}}2\bar{w}(\bx,\by)\left(\CP_i \bu^t(\bx)-\CP_i \bu^t(\by)\right)\right.\\ \nonumber
  &+\left.\mu\sum_{\by \in \Omega_i^t}\bar{w}(\bx,\by)\left(\CP_i \bu^t(\bx)-\CP_i \bu^t(\by)\right)\right]\\ \label{eq:el-original}
  &+ \lambda I_{\Omega^t}\left(\bu^t-\bb^t\right), \quad \forall \bx \in \bar{\Omega}
\end{align}
where $\mu = 1/r-1$, $\CP_i^*$ is the adjoint operator of $\CP_i$, $I_{\Omega^t}$ is the projection operator that sets  $\bu^t(\bx)$ to  zero for  $\bx \notin \Omega^t$. We use the notation $\bx_{\widehat{j}}$ to denote the $j$-th component (in the spatial domain) after $\bx$ in a patch. It is easy to verify that $\CP_i \bu^t(\bx) = \bu^t(\bx_{\widehat{i-1}})$, and $\CP_i^* \bu^t(\bx) = \bu^t(\bx_{\widehat{1-i}})$. Following the analysis similar to \cite{ldmm_scientific}, we can rewrite (\ref{eq:el-original}) as
\begin{align}
\nonumber
  0  =  &\mu I_{\Omega^t}\left[ \sum_{\by \in \bar{\Omega}}\sum_{i=1}^{d_s}\bar{w}(\bx_{\widehat{1-i}},\by_{\widehat{1-i}})\left(\bu^t(\bx)-\bu^t(\by)\right) \right]\\\nonumber
  &+ \sum_{i=1}^{d_s}\left[\sum_{\by \in \bar{\Omega}}2\bar{w}(\bx_{\widehat{1-i}},\by_{\widehat{1-i}})\left(\bu^t(\bx)-\bu^t(\by)\right)\right.\\ \nonumber
  &+ \left.\mu\sum_{\by \in \Omega^t}\bar{w}(\bx_{\widehat{1-i}},\by_{\widehat{1-i}})\left(\bu^t(\bx)- \bu^t(\by)\right)\right]\\  \label{eq:el}
  &+\lambda I_{\Omega^t}\left(\bu^t-\bb^t\right). \quad \forall \bx \in \bar{\Omega}
\end{align}

After setting $\tilde{w}(\bx,\by)= \sum_{i=1}^{d_s} \bar{w}(\bx_{\widehat{1-i}},\by_{\widehat{1-i}})$, (\ref{eq:el}) is equivalent to
\begin{align}\nonumber
  0 = &2\sum_{y\in \bar{\Omega}}\tilde{w}(\bx,\by)\left(\bu^t(\bx)-\bu^t(\by)\right)+\lambda I_{\Omega^t}\left(\bu^t-\bb^t\right). \\\nonumber
  + &\mu I_{\Omega^t}\left[\sum_{\by\in\bar{\Omega}}\tilde{w}(\bx,\by)\left(\bu^t(\bx)-\bu^t(\by)\right)\right]\\ \label{eq:linear-system}
  +&\mu\sum_{y\in\Omega^t}\tilde{w}(\bx,\by)\left(\bu^t(\bx)-\bu^t(\by)\right) ,\quad \forall \bx \in \bar{\Omega}
\end{align}

Note that (\ref{eq:linear-system}) is a linear system for $\bu^t$ in $\R^{m n}$, but unlike \cite{ldmm_scientific}, the coefficient matrix is not symmetric because of the projection operator $I_{\Omega^t}$. In our numerical experiments, we always truncate the similarity matrix $\bar{w}(\bx,\by)$ to 20 nearest neighbors. Therefore, (\ref{eq:linear-system}) is a sparse linear system and can be solved by the generalized minimal residual method (GMRES). The proposed algorithm for HSI reconstruction is summarized in Algorithm~\ref{alg}.

\begin{algorithm}
\floatname{algorithm}{Algorithm}
\caption{Scalable LDMM for HSI reconstruction}
\label{alg}
\begin{algorithmic}
\REQUIRE A noisy and incomplete observation $\bb$ of an unknown hyperspectral image $\bu \in \R^{m \times n \times B}$. For every spectral band $t \in [B]$, $\bu$ is only partially observed on a random subset $\Omega^t$ of $\bar{\Omega}=[m]\times[n]$.
\ENSURE  Reconstructed HSI $\bu$.
\STATE Initial guess $\bu^{(0)}$.
\WHILE {not converge}
\STATE 1. Extract the patch set $\CP\bu^{(k)}$ from  $\bu^{(k)}$.
\STATE 2. Compute the similarity matrix on the spatial domain 
$$\overline{w}(\bx,\by)=w(\CP \bu^{(k)}(\bx),\CP \bu^{(k)}(\by)),\quad \bx,\by\in \overline{\Omega}.$$ 
\STATE 3. Assemble the new similarity matrix
$$\tilde{w}(\bx,\by)=\sum_{i=1}^{d_s}\bar{w}(\bx_{\widehat{1-i}},\by_{\widehat{1-i}})$$
\STATE 4. For every spectral band $t$, Update $(\bu^{t})^{(k+1)}$ as the solution of (\ref{eq:linear-system}) using GMRES.
\STATE 5. $k\leftarrow k+1$.
\ENDWHILE
\STATE $\bu=\bu^{(k)}$.
\end{algorithmic}
\end{algorithm}

\section{Numerical Experiments}
\subsection{Experimental Setup}

In this section, we present the numerical results on the following datasets: Pavia University (PU), Pavia Center (PC), Indian Pine (IP), and San Diego Airport (SDA). All images have been cropped in the spatial dimension to $200\times 200$ for easy comparison. The objective of the experiment is to reconstruct the original HSI from 5\% random subsample (10\% random subsample for noisy data).

Empirically, we found out that it is easier for LDMM to converge if a reasonable initialization is provided. In our experiments, we always use the result of the low-rank matrix completion algorithm APG \cite{lr} as an initialization, and run three iterations of manifold update for LDMM. The peak signal-to-noise ratio, $\text{PSNR} = 10\log_{10}\left(\|\bu^*\|_{\infty}/\text{MSE}\right)$, is used to evaluate the reconstruction, where $\bu^*$ is the ground truth, and MSE is the mean squared error. All experiments were run on a Linux machine with 8 Intel core i7-7820X 3.6 GHz CPUs and 64 GB of RAM. All codes and datasets are available for download at \url{http://www.math.duke.edu/~zhu/software.html}.

\begin{table}[H]
  \centering
  \begin{tabular}{C{.7cm}|C{.9cm}|C{.75cm}|C{.9cm}|C{.75cm}|C{.9cm}|C{.75cm}}
    \hline\hline
    & \multicolumn{2}{c|}{APG} & \multicolumn{2}{c|}{LDMM1} & \multicolumn{2}{c}{LDMM2}\\\cline{2-7}
    &PSNR&time&PSNR&time&PSNR&time\\ \hline
    IP & 26.80 & 13 s & 32.09 & 8 s & \textbf{34.08} & 22 s\\\hline
    PC & 32.61 & 17 s & \textbf{34.54} & 11 s & 34.25 & 31 s\\\hline
    PU & 31.51 & 13 s & 33.38 & 11 s & \textbf{33.66} & 29 s\\\hline
    SDA & 32.43 & 23 s & 40.33 & 16 s & \textbf{44.21} & 46 s\\
    \hline\hline
  \end{tabular}
  \vspace{-2mm}
  \caption{Reconstruction of the HSIs from their noise-free 5\% subsamples. LDMM1 (LDMM2) stands for LDMM with spatial patch size of $1\times 1$ ($2\times 2$). The reported time of LDMM does not include that of the AGP initialization.}
  \label{tab:result_noisefree}
\end{table}

\begin{figure}[H]
  \centering
  \begin{tabular}{cc}
    Original (Band 33) & 5\% subsample\\
    \includegraphics[width=0.4\linewidth]{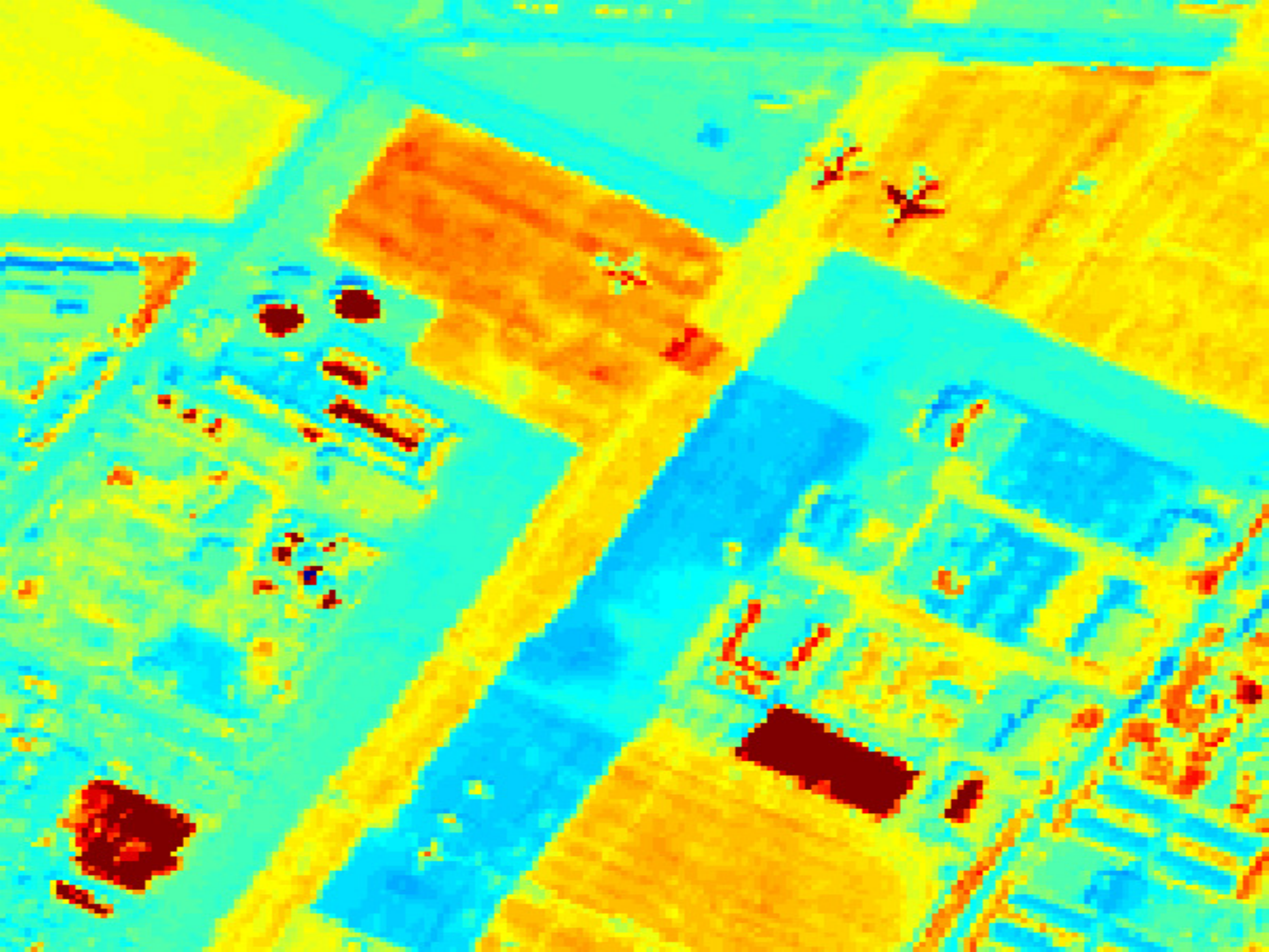}&
    \includegraphics[width=0.4\linewidth]{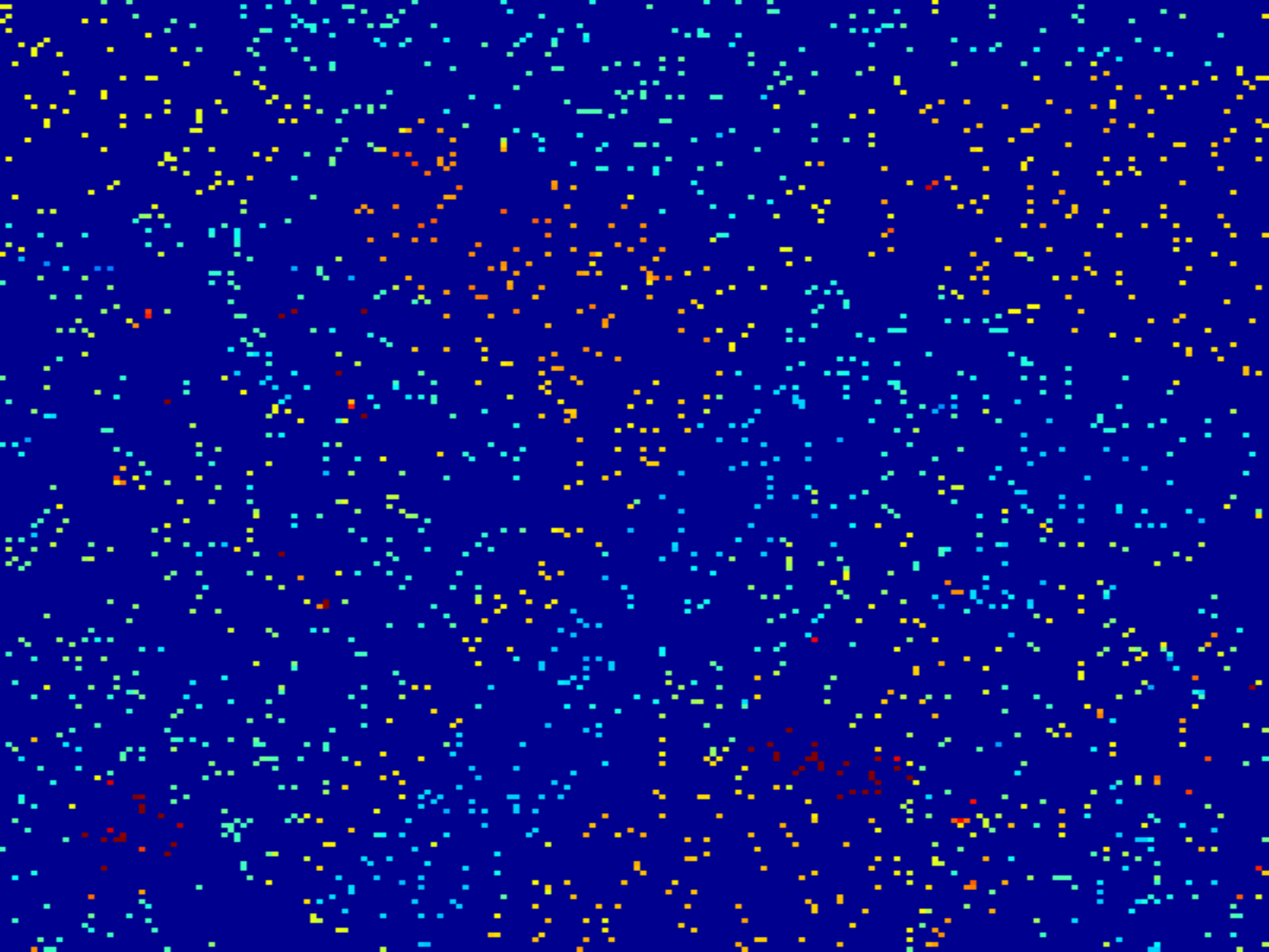}\\
    APG (PSNR = 32.43) & Error\\
    \includegraphics[width=0.4\linewidth]{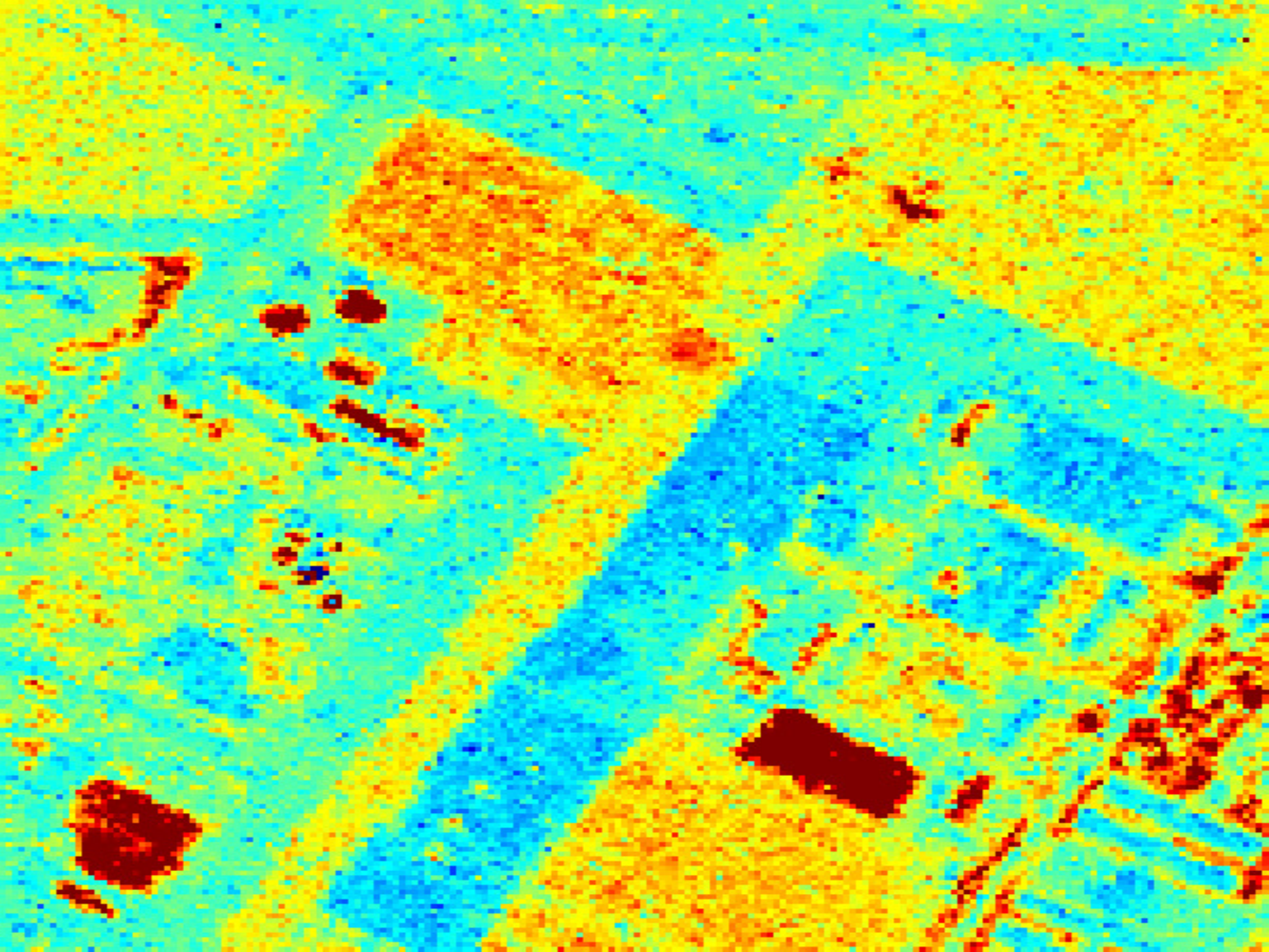}&
    \includegraphics[width=0.4\linewidth]{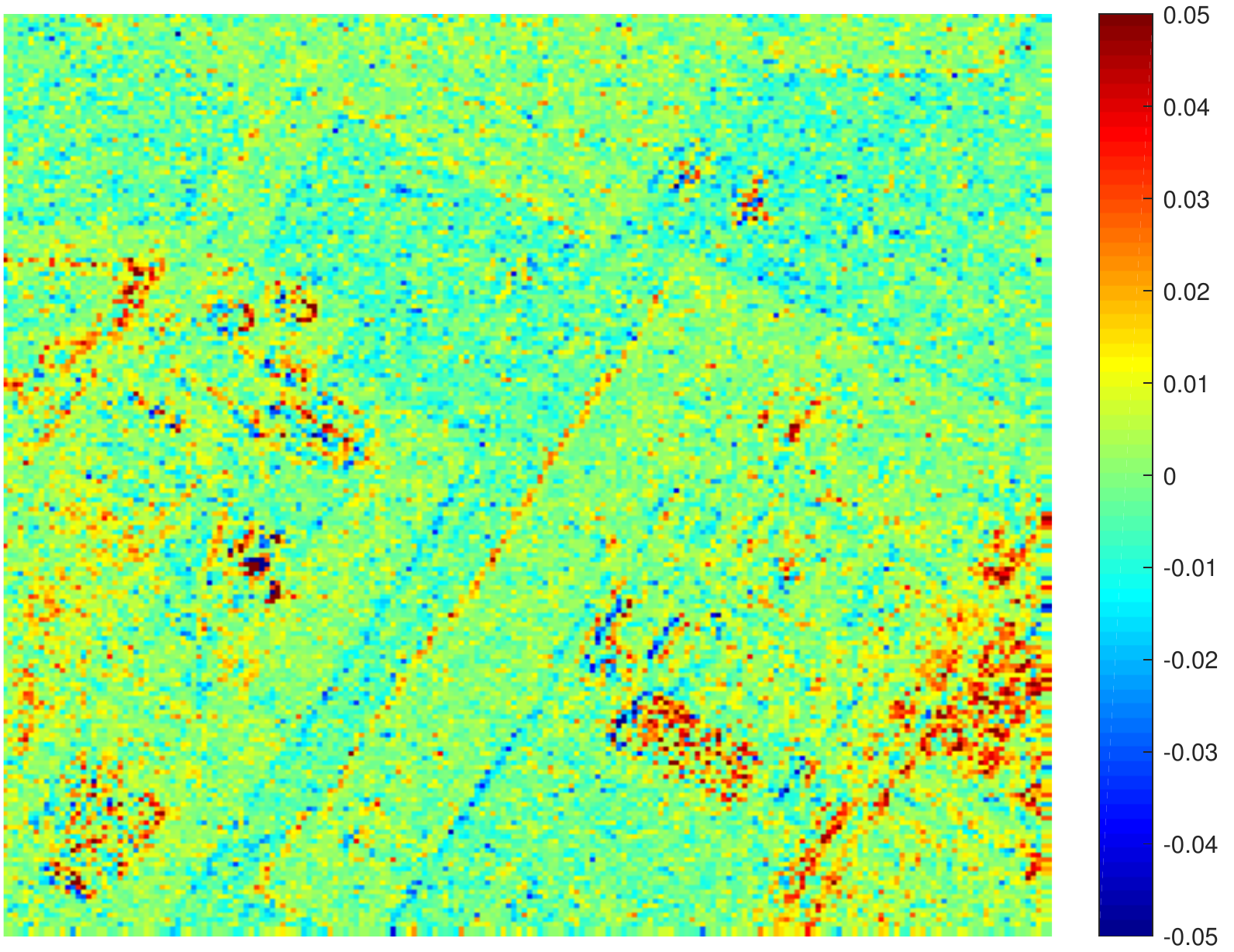}\\
    LDMM (PSNR = 44.21) & Error\\
    \includegraphics[width=0.4\linewidth]{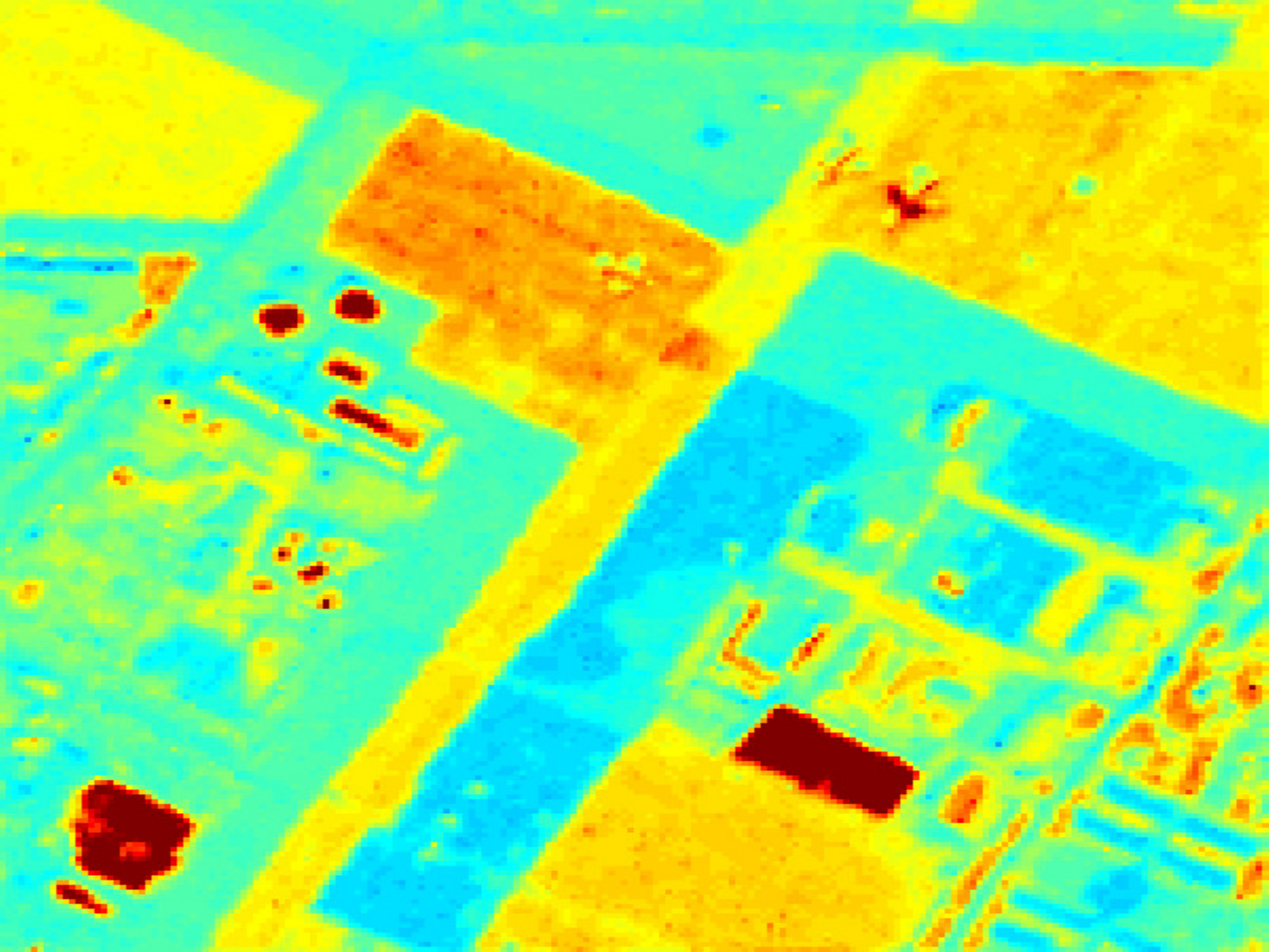}&
    \includegraphics[width=0.4\linewidth]{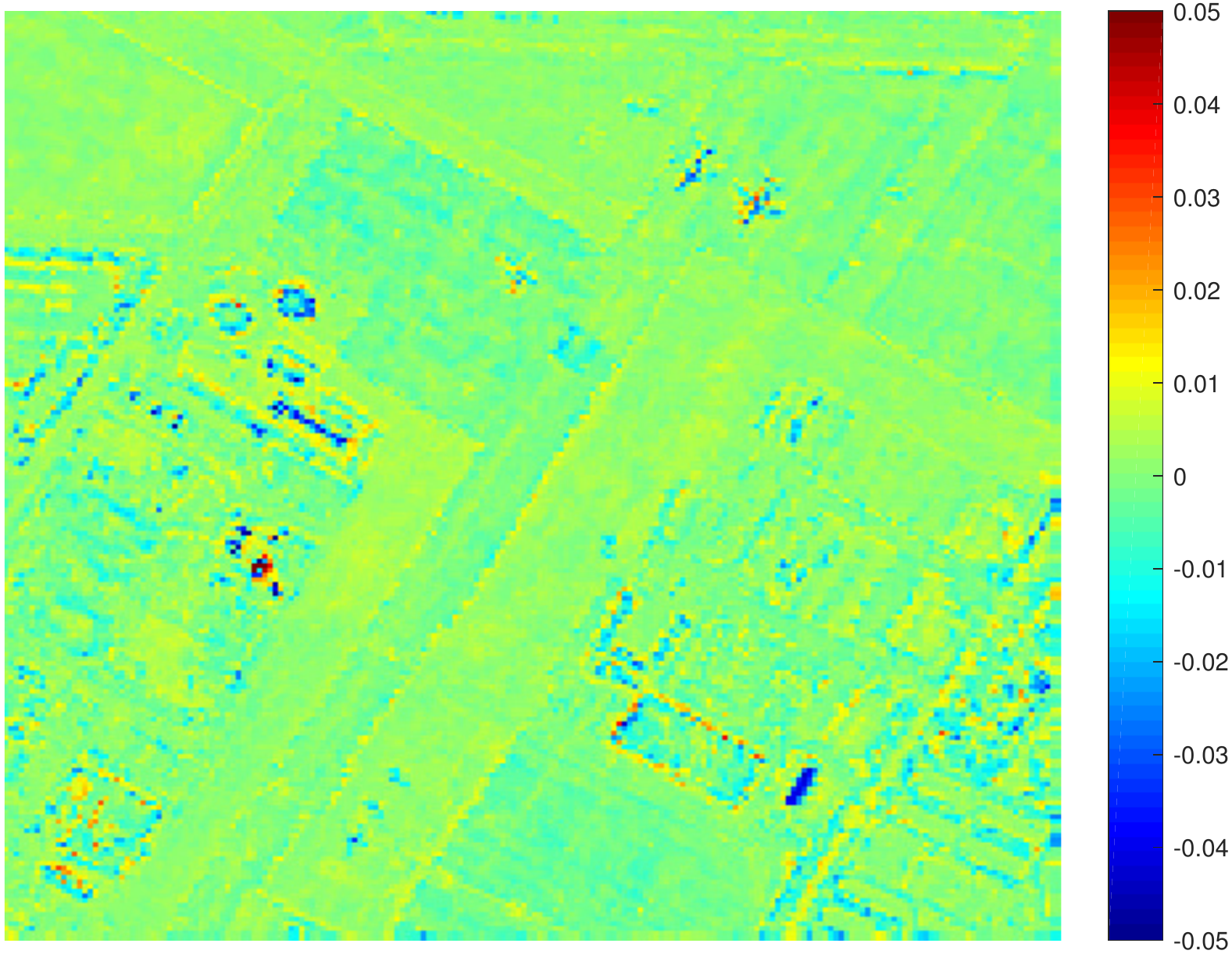}
  \end{tabular}
  \caption{Reconstruction of SDA from 5\% noise-free subsample. Note that the error is displayed with a scale 1/20 of the original data to visually amplify the difference.}
  \label{fig:noise-free}
\end{figure}

\subsection{Reconstruction from noise-free subsample}

We first present the results of the reconstruction of HSI from 5\% noise-free random subsample. Table~\ref{tab:result_noisefree} displays the computational time and accuracy of the low-rank matrix completion (APG) initialization and LDMM with different spatial patch sizes ($1\times 1$ and $2\times 2$). It can be observed that LDMM significantly improves the accuracy of APG with comparable extra computational time. Figure~\ref{fig:noise-free} provides a visual illustration of the results. Because of the limited space, we only present the reconstruction of SDA on one spectral band.

\subsection{Reconstruction from noisy subsample}

\begin{figure}
  \centering
  \begin{tabular}{cc}
    Original (Band 38) & Noise added\\
    \includegraphics[width=0.45\linewidth]{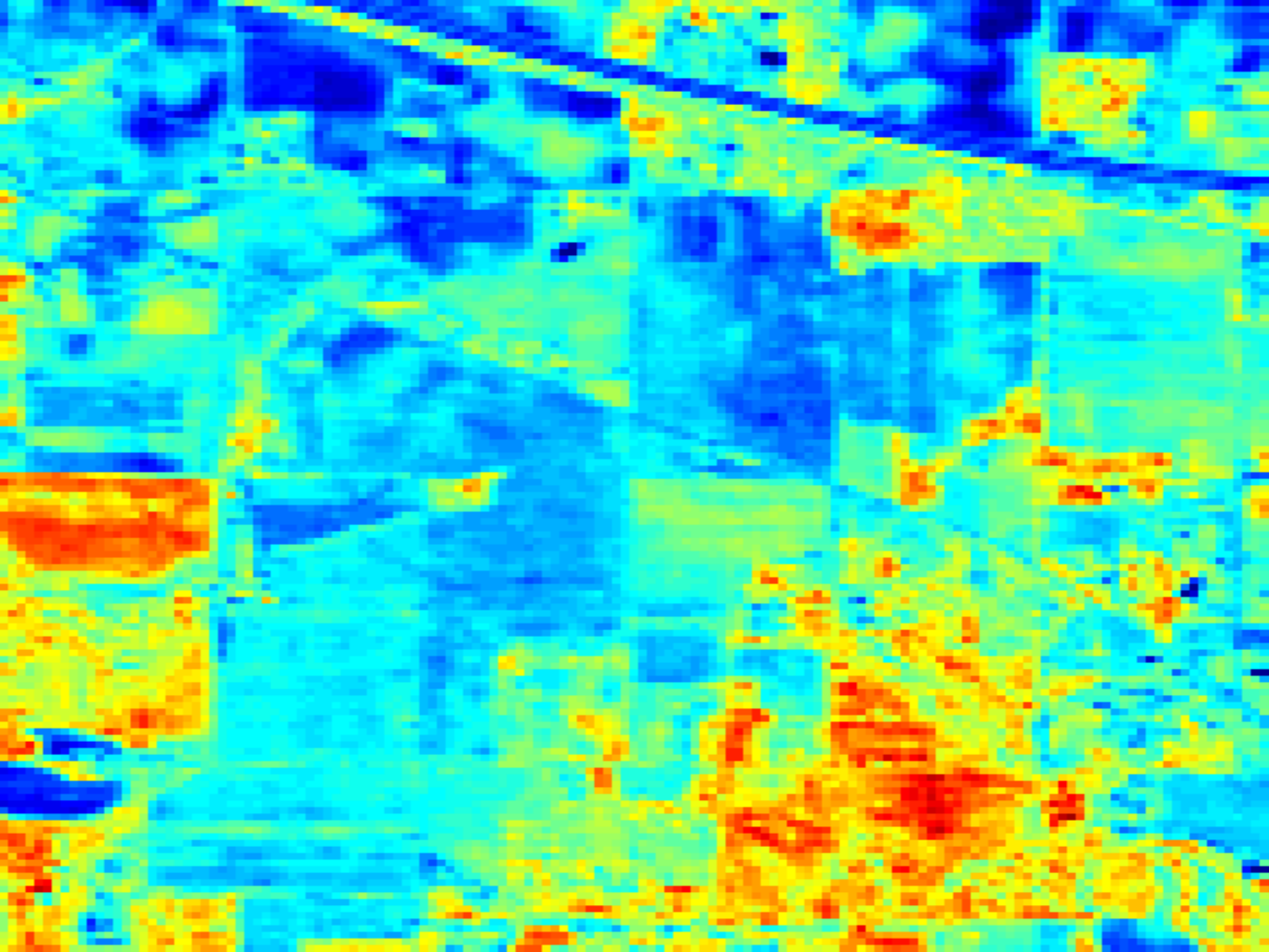}&
    \includegraphics[width=0.45\linewidth]{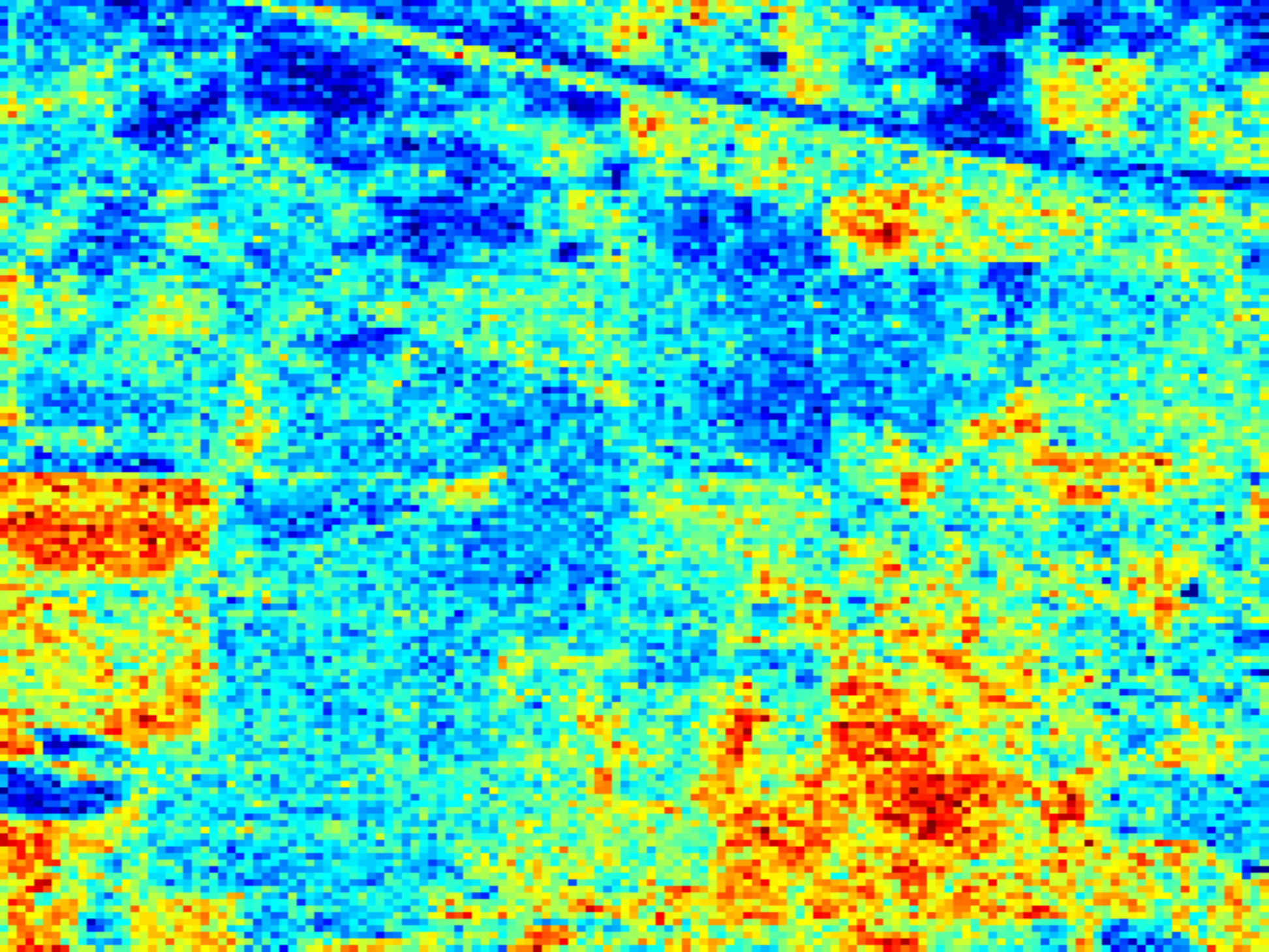}\\
    10\% noisy subsample & LDMM\\
    \includegraphics[width=0.45\linewidth]{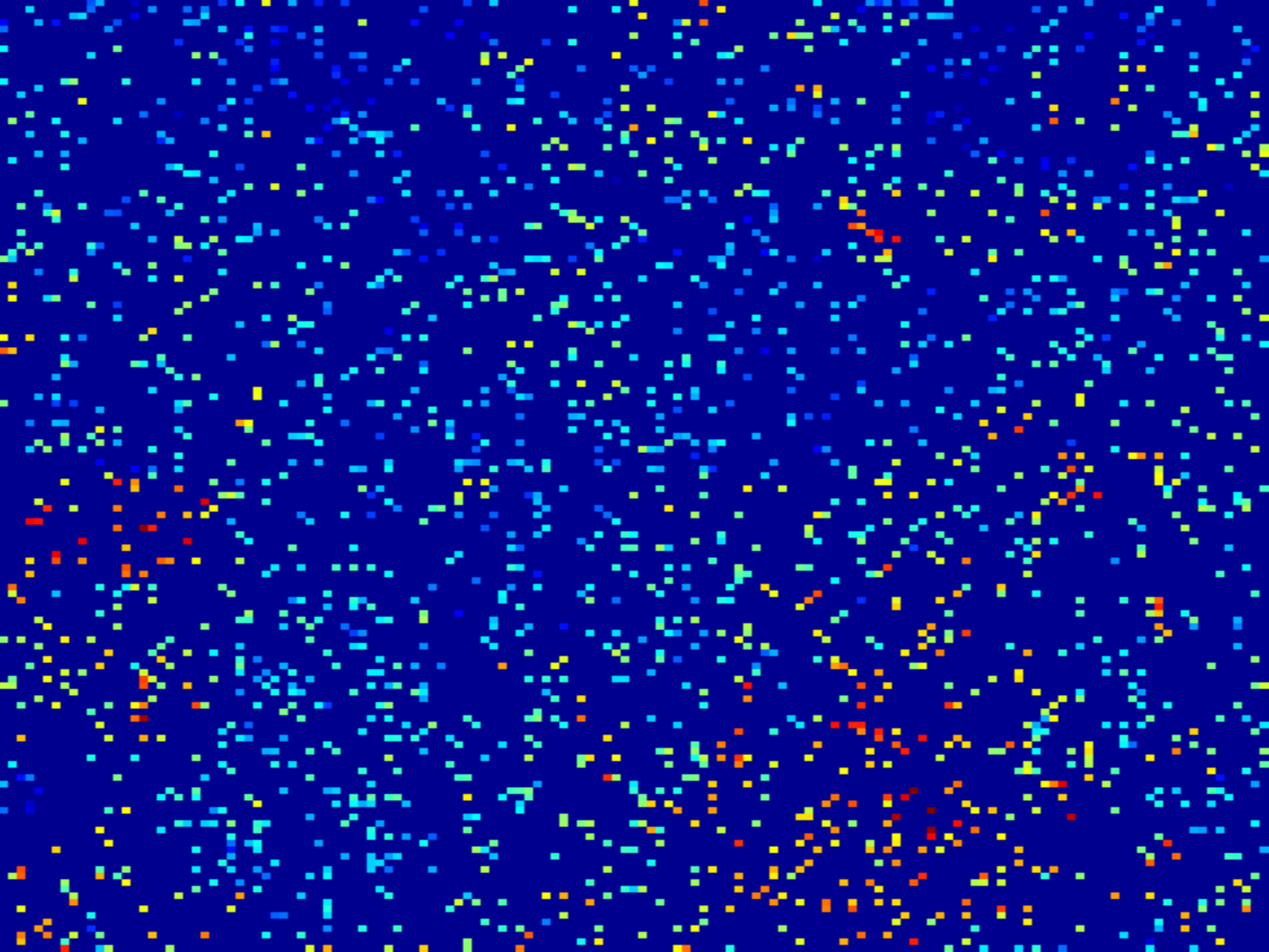}&
    \includegraphics[width=0.45\linewidth]{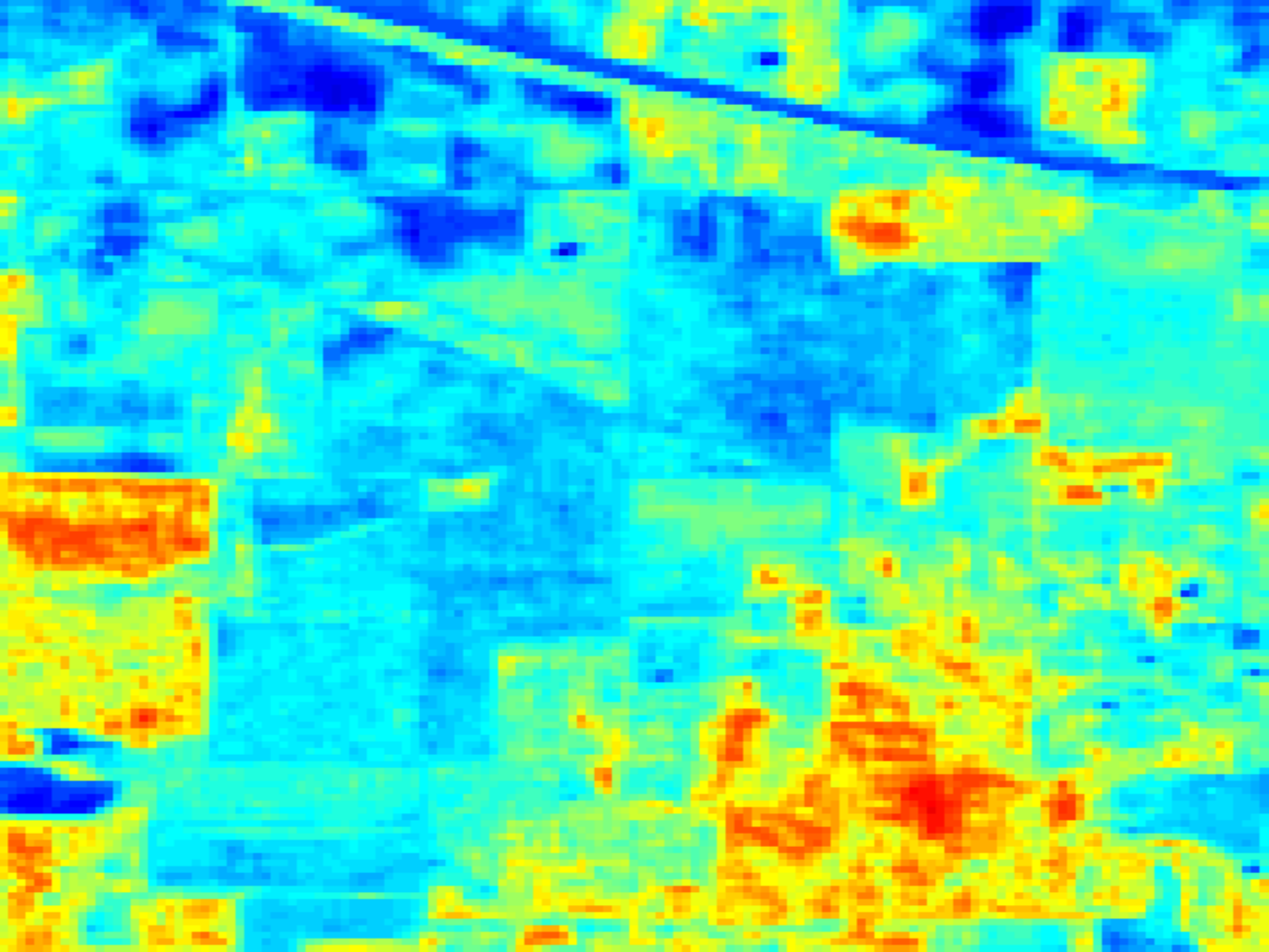}
  \end{tabular}
  \caption{Reconstruction of Indian Pine from 10\% noisy subsample.}
  \label{fig:noisy}
\end{figure}

We then show the results of the reconstruction of HSI from 10\% noisy subsample. A gaussian noise with a standard deviation of $0.05$ is added to the original image, and then we remove 90\% of the voxels from the data cube. The accuracy and computational time is reported in Table~\ref{tab:result_noisy}. Note that LDMM with $2\times 2$ patches typically produce better results than that with $1\times 1$ patches because of the presence of noise. A visual demonstration of the reconstruction is displayed in Figure~\ref{fig:noisy}.

\begin{table}
  \centering
  \begin{tabular}{C{.7cm}|C{.9cm}|C{.75cm}|C{.9cm}|C{.75cm}|C{.9cm}|C{.75cm}}
    \hline\hline
    & \multicolumn{2}{c|}{APG} & \multicolumn{2}{c|}{LDMM1} & \multicolumn{2}{c}{LDMM2}\\\cline{2-7}
    &PSNR&time&PSNR&time&PSNR&time\\ \hline
    IP & 31.56 & 18 s & \textbf{34.03} & 54 s & 34.02 & 56 s\\\hline
    PC & 30.22 & 47 s & 30.55 & 82 s & \textbf{31.61} & 82 s\\\hline
    PU & 29.88 & 38 s & 30.26 & 77 s & \textbf{31.40} & 86 s\\\hline
    SDA & 33.90 & 69 s & 39.17 & 186 s & \textbf{41.31} & 231 s\\
    \hline\hline
  \end{tabular}
  \vspace{-2mm}
  \caption{Reconstruction of the noisy HSIs from their 10\% subsamples. LDMM1 (LDMM2) stands for LDMM with spatial patch size of $1\times 1$ ($2\times 2$). The reported time of LDMM does not include that of the AGP initialization.}
  \label{tab:result_noisy}
\end{table}

\section{Conclusion}
We propose the scalable low dimensional manifold model for the reconstruction of hyperspectral images from noisy and incomplete observations with a significant number of missing voxels. The dimension of the patch manifold is directly used as a regularizer, and the same similarity matrix is shared across all spectral bands, which significantly reduces the computational burden. Numerical experiments show that the proposed algorithm is an accurate and efficient means for HSI reconstruction.

\bibliographystyle{IEEEbib}
\bibliography{ldmm_hsi}

\end{document}